# Towards a Benchmark for Scientific Understanding in Humans and Machines

**Kristian Gonzalez Barman[a], Sascha Caron[b,c], Tom Claassen[d], Henk de Regt[a]**

[a] *Institute for Science in Society, Faculty of Science, Radboud University, the Netherlands.*

[b] *High Energy Physics, Faculty of Science, Radboud University, the Netherlands.*

[c] *Nikhef, Science Park 105, 1098 XG Amsterdam, the Netherlands.*

[d] *Institute for Computing and Information Sciences, Faculty of Science, Radboud University, the Netherlands.*

E-mail: kristian@gonzalezbarman@ru.nl , scaron@nikhef.nl , tomc@cs.ru.nl , henk.deregt@ru.nl

## Abstract

Scientific understanding is a fundamental goal of science, allowing us to explain the world. There is currently no good way to measure the scientific understanding of agents, whether these be humans or Artificial Intelligence systems. Without a clear benchmark, it is challenging to evaluate and compare different levels of and approaches to scientific understanding. In this Roadmap, we propose a framework to create a benchmark for scientific understanding, utilizing tools from philosophy of science. We adopt a behavioral notion according to which genuine understanding should be recognized as an ability to perform certain tasks. We extend this notion by considering a set of questions that can gauge different levels of scientific understanding, covering information retrieval, the capability to arrange information to produce an explanation, and the ability to infer how things would be different under different circumstances. The Scientific Understanding Benchmark (SUB), which is formed by a set of these tests, allows for the evaluation and comparison of different approaches. Benchmarking plays a crucial role in establishing trust, ensuring quality control, and providing a basis for performance evaluation. By aligning machine and human scientific understanding we can improve their utility, ultimately advancing scientific understanding and helping to discover new insights within machines.

# Introduction

This Roadmap presents a framework for measuring scientific understanding in agents, including humans, machine learning models, and model-augmented humans[1,2]. Current benchmarks in Machine Learning measure a variety of capabilities[3,4]. For example, the Winograd Schema Challenge[5] (WSC) and the General Language Understanding Evaluation[6] (GLUE) measure linguistic understanding, while BIGBench[7] measures proficiency at several tasks such as simple logic problems or guessing a chess move. However, despite their need and importance, there are currently no benchmarks that measure the degree of scientific understanding. To address this gap, we provide definitions of scientific understanding, a framework for how it can be measured, and discuss potential use cases such as discovering new insights within machines.

The primary objective of this Roadmap is to provide a foundation for benchmarking and measuring the scientific understanding of various agents (including Large Language Models[8,9] and Question Answering Machines[10]). Although our focus is on scientific understanding [11,12] in the natural sciences (e.g., physics), we anticipate that our framework can be applied to numerous other scientific disciplines. We break with the traditional view [13,14,15,16,17] by arguing that understanding should be conceptualized in terms of abilities rather than internal mechanics[18,19] or representations[14,20]. Specifically, we contend that scientific understanding is a skill-based capability that relies on an agent's ability to perform specific actions, rather than a subjective mental state. This perspective separates the subjective 'feeling' of understanding from genuine understanding, indicating that psychological states are neither sufficient nor necessary to establish understanding[21]. We preliminarily define [22,12] scientific understanding as the ability to provide explanations within a theoretical framework that is intelligible to the agent, which involves the ability to derive qualitative results, answer questions, solve problems properly, and extend knowledge to other domains or levels of abstraction. We argue that various degrees of scientific understanding can be measured by different levels of ability, such as having access to relevant information, the ability to provide explanations, and the ability to establish counterfactual inferences. These different levels can be quantitatively evaluated using what-, why-, and w-questions (see below).

Our framework enables creating specific tests, which can be used for benchmarking models, measuring student understanding, and evaluating teaching abilities, or training a machine learning model. We provide guidelines for researchers, including different testing interfaces. We then propose the creation of a benchmark for scientific understanding. This benchmark is an important first step towards assessing machine understanding, where aligning machine understanding with human understanding can help in research tasks, such as hypothesis creation and information retrieval and summarization.

It should be noted that the tests stemming from our framework differ significantly from the Turing Test[23,24]. Their focus is not on evaluating general intelligence but rather scientific understanding. The score one obtains in scientific understanding has no meaning as to whether an agent has AGI status or other abilities beyond scientific understanding. Similarly, passing the Turing Test does not necessarily mean an agent would score high on the scientific understanding test.

We compare our framework to a recently proposed[25] student-teacher interaction as a test of scientific understanding in machines and show how incorporating elements of our framework could improve this test and offer a quantifiable measure of transfer of understanding between agents. While their approach to testing focuses on evaluating new understanding, our framework is capable of testing existing understanding as well as new understanding, where new understanding might involve increasing the understanding of phenomena (e.g., by providing deeper explanations) or discovering new phenomena. Finally, we discuss the abilities and limitations of our framework considering popular LLM implementations.

Our contributions can be summarized in the following points:

- Scientific understanding is an ability and should therefore be measured in terms of behavioral competence (i.e., actions).
- A framework that provides a basis for developing tests to measure scientific understanding both in human and artificial agents. The framework can be used for benchmarking models, assessing student understanding, and training machine learning models.
- Guidelines for implementing tests to measure the scientific understanding of Large Language Models (LLMs) together with a call for scientific communities to systematically engage in benchmarking question-answering machines, to foster specific developments such as testing new scientific understanding (i.e., discovery of new insights unknown to humankind) within machines.
- A discussion on how to test whether a machine has transferred scientific understanding to another agent, building on a recently proposed account.

## Scientific understanding as an ability

Scientific understanding is traditionally viewed as an internal mental state or representation possessed by an agent, such as a human scientist.[26] [27] This perspective focuses on the subjective and internal aspects of understanding, such as mental representations or the "feeling" of understanding, rather than the observable aspects of the agent's abilities and actions.

Floridi[28] appears to endorse this traditional view, expressing skepticism about the capacity of Large Language Models (LLMs) to achieve any degree of understanding, as they do not reason or resemble the cognitive processes present in animal or human brains. Critics of LLMs often argue that agents must have the same underlying mechanisms as humans to understand. However, this argument is insufficient, as it is unclear why understanding could not be realizable through different mechanisms. Moreover, human understanding is rarely assessed based on underlying mechanisms, but rather on behavioral observations, such as answering questions in an exam.

Here we propose a re-evaluation of scientific understanding, arguing that it should be assessed based on an agent's ability to perform certain tasks, rather than the underlying mechanisms involved in those tasks. This means that artificial agents, including Large Language Models (LLMs) should not be dismissed as incapable of scientific understanding simply because they "guess the next word" or are "stochastic parrots" [29]. By integrating philosophical accounts with empirical and theoretical bases, we critically evaluate machine understanding, contending that scientific understanding is an ability that should be assessed based on behavior, rather than mental representations, internal architecture, consciousness, or other similar factors.

We maintain that evaluating all agents, including LLMs, should follow the same principles used to assess human scientific understanding—based on their abilities to perform relevant tasks. We are not the only ones who relate understanding to an ability [30][25][20]. For example, Tamir and Shech[20] have argued that practical abilities (such as reliable and robust task performance) can be seen as key factors indicative of understanding in the context of deep learning. While we think this is a good start, we argue that a more comprehensive and rigorous evaluation of understanding as an ability is needed. That is precisely what this Roadmap aims to achieve.

## A Framework for Scientific Understanding

Our starting point is de Regt's[12] account of scientific understanding, on which understanding a phenomenon boils down to having an adequate explanation of the phenomenon within the right theoretical scaffolding. The formal criterion is the following[31] (Criterion for Understanding a Phenomenon):

> **CUP:** A phenomenon P is understood scientifically if and only if there is an explanation of P that is based on an intelligible theory T and conforms to the basic epistemic values of empirical adequacy and internal consistency.

Note that the use of a biconditional indicates that this is a necessary and sufficient condition. The explanation must be based on an intelligible theory. A test for intelligibility can described by the following criterion[32]:

> **CIT$_1$:** A scientific theory $T$ (in one or more of its representations) is intelligible for scientists (in context $C$) if they can recognize qualitatively characteristic consequences of $T$ without performing exact calculations.

In this case, the condition is sufficient but not necessary. This is also why there is a subscript 1 in CIT$_1$, since there might be other criteria for intelligibility. CIT$_1$ holds primarily for theories with a mathematical formulation, such as in physics. For other types of theories, other conditions might hold. The key aspect of this condition is the ability to derive (qualitative) consequences.

To elaborate this conception of scientific understanding, we modify the definition by shifting the focus from the phenomenon being understood to the conditions required for an agent to understand. We develop these conditions into having access to information, having explanatory abilities (since they might not coincide), and reformulate the ability to recognize qualitatively characteristic consequences in terms of counterfactual inferences. Moreover, we refine the definition by emphasizing the importance of measuring understanding instead of relying on strict necessary and sufficient conditions. Doing so acknowledges that understanding is not a binary affair but rather a matter of degree. These extensions result in the following **general framework for an agent's scientific understanding** (**A**gent-**U**nderstands-**P**henomenon):

> **AUP**: The degree to which agent A scientifically understands phenomenon P can be determined by assessing the extent to which (i) A has a sufficiently complete representation of P; (ii) A can generate internally consistent and empirically adequate explanations of P; (iii) A can establish a broad range of relevant, correct counterfactual inferences regarding P.

**AUP** is our framework for establishing the degree of scientific understanding of a phenomenon (by an agent). This framework can be instantiated in different ways (e.g., there might be several ways of establishing whether A has a sufficiently complete representation of P). One implementation would be $AUP_1$:

> $AUP_1$(i-iii) can be measured, given a certain context (series of prompts) via what-, why, and w-questions respectively.

These questions are prompt specific, where the context (i.e., initial prompting to provide necessary information) and the ordering of questions makes a difference. This feature makes the application $AUP_1$ dynamic.

The first level ('i') of **AUP** requires having access to sufficient relevant information about P. This access involves the capacity to retrieve information from relevant sources (such as memories, encodings/embeddings, databases, or the internet). We argue that this can be measured by the ability to provide correct answers to 'what-questions' (see section below).

The second level ('ii') refers to the capability of arranging information to produce an explanation of P. The ability to generate a well-constructed explanation surpasses simple information retrieval, requiring a deeper level of understanding[33]. We argue that the ability to provide explanations can be evaluated through answers to why-questions (see section below).

The third level ('iii') is the ability to infer how P would have been (or would be) different under different circumstances; namely, the ability to draw counterfactual inferences. This ability requires being able to properly *use* a (good) explanation[34][35][36]. We argue that this ability can be measured via answers to w-questions (see section below). Answers to w-questions require more than simply having an explanation, they require having a good explanation and knowing how to use it (e.g., knowing when the explanation is applicable, what the boundary conditions are, etc.). W-questions assess competency at establishing counterfactual inferences concerning a phenomenon and can be linked to an agents' breadth and depth of understanding [37].

# Test Questions

## What-questions

What-questions ask for descriptive knowledge about an object or phenomenon[38][39]. Answering such questions requires having access to information, whether from memory or external sources (books, servers, etc.). What-questions can ask for values, dimensions, or names, among other things. For example, what is the charge of the electron? The ability to answer what-questions correctly is a necessary but not a sufficient condition for understanding, as it only involves the retrieval of information and not the ability to use said information for higher-level tasks (see section above).

### Why-questions (explanation-seeking questions)

Answering why-questions [40] [41] [42] which inquire about facts or phenomena requires providing an explanation. It is for this reason that answering them correctly is highly indicative of scientific understanding. Why-questions can be divided into (at least) three types:

1. Questions of singular facts: 'Why is it the case that A?' ('Why is charge conserved?')
2. Contrastive questions[43]: 'Why A rather than B' ('Why did Patient A rather than Patient B get better with treatment T?', 'Why did Patient P get better using treatment A rather than treatment B?')
3. Resemblance questions[44]: 'Why do A and B share C' ('Why do both hedgehogs and bears hibernate?')

Answering why-questions requires articulating information in a way that is sensitive to context and explanatory aims[43]. Using a variety of why-questions with answers not easily found (e.g., by choosing different foils or contrast classes in the case of contrastive explanations) can help ensure an explanatory ability that is not simply due to memorization or accessing the internet.

### W-questions (counterfactual inferences)

W-questions refer to what-if-things-had-been-different questions[34] and what-would-happen-if questions[45]. These questions explore alternative scenarios and potential outcomes based on a hypothetical change in circumstances. Answers to what-if-things-had-been-different questions enable us to see what the outcome of some state of affairs would have been if initial conditions had been different. Answers to what-would-happen-if questions can be thought of as a prediction that involves some sort of manipulation or intervention on a system[45].

Answering these two types of questions can be thought of as backward-looking and forward-looking counterfactual inferences. In both cases, answering these questions involves postulating hypothetical scenarios about what would occur under a specific set of circumstances. It is this feature that we are interested in, since the ability to adequately derive these scenarios requires understanding. Similarly, there is a strong link between the quality of an explanation and the counterfactual inferences it affords [34] [46] [47]. We therefore contend that the range of counterfactual inferences an agent can articulate is strongly related to the level of understanding [48]. Counterfactual inferences can be more or less general, where one can distinguish between parameterized and exogenous variable counterfactual inferences[49] [50]. Parameterized counterfactual inferences involve changing specific variables within a model, such as "What would happen to the period of this spring-mass system if we changed the spring constant?". Exogenous variable counterfactual inferences involve changing external variables that impact the system, such as "What would happen to this spring-mass system if the spring breaks?".

# From a general framework to specific tests

In this section we discuss how to operationalize the framework described in the previous section into concrete tests to measure scientific understanding. Understanding can be of a concrete phenomenon (e.g., a pendulum of 5m length, 2kg weight, etc.) or of a general phenomenon (e.g., pendulums in general). Tests can be devised for both specific and general phenomena, and the level of generality can be increased by asking higher-level w-questions, such as those related to changing exogenous variables.

## How to score an agent?

The level of scientific understanding of an agent can be thought of as a gradient between complete lack of understanding to an ever-increasing level (See figure 1). The agent's score would depend on the number of correct answers, with varying weights assigned to different questions. This test can provide a specific score for the scientific understanding of an agent or compare two agents. We can establish different thresholds depending on the context.

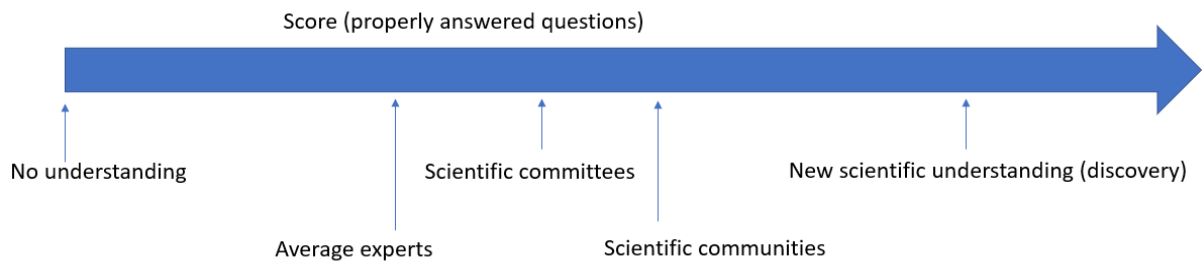

*Figure 1. Scientific understanding can be categorized into various levels based on the number of questions answered. An agent possessing the ability to answer all the questions posed by scientific communities, including those for which we do not yet have answers, indicates a higher level of (new) scientific understanding.*

Each individual test should contain a sufficiently diverse and representative set of questions that can capture enough details of the properties, attributes, and elements of the phenomenon. This implies that question generation should be conducted by the experts of each community. We provide some guidelines for this below. In some cases, it could be possible to train language models to produce questions as well [51] [52] [53].

## Guidelines for testing

There is a need for guidelines to establish a standardized and reliable approach to testing which ensures accurate and consistent results. In developing a comprehensive and reliable test for AI agents, it is crucial to define the test scope and purpose, ensuring it is tailored to the agent and includes a variety of difficulty levels. To achieve consistent, repeatable scoring, diverse question formats should be employed, with multiple testing instances conducted for robust evaluation. Crafting comprehensive, varied, and representative questions is essential, using concise and unambiguous language to prevent confusion. To maintain test integrity, limit answer accessibility on the internet and other public repositories. Finally, centralized storage of tests for easy review, enabling them to serve as part of the SUB benchmark. Additional guidelines for good testing and evaluation can also be implemented [54] [55] [56] [57].

It is important to note that this test should be conducted through an interface, which may need to be tailored for certain agents. Traditional testing methods such as multiple-choice tests can be used for humans and model-augmented humans, while an interface that allows for context encoding and some form of chat-like interface (e.g., ChatGPT [58]) is needed for LLMs.

## The Scientific Understanding Benchmark (SUB)

We propose two things. First, a call to communities to create tests for scientific understanding to benchmark different AI models. Benchmarking plays a crucial role in establishing trust in the reliability of models, ensuring quality control, and providing a basis for performance evaluation. Given the current situation in AI it is thereby of high societal relevance.[59] Second, the bringing together of tests developed by different communities into a broad benchmark, which we call the Scientific Understanding Benchmark (**SUB**). This should be an open project supervised by an independent community of experts that, among other criteria, sets high standards for scientific correctness.

We firmly believe that the **SUB** will have a positive impact on the usefulness utility, confidence, and controllability of AI in scientific research and expect it to advance scientific understanding, facilitate stakeholder alignment, and enable new discoveries.

## Scientific understanding transfer

Krenn et al.[25] closely follow an earlier version of de Regt's (2017) account, namely, de Regt and Dieks (2005)[22]. Based on **CIT** (see the section on Scientific Understanding) they formulate a parallel condition replacing the scientist(s) with an AI. Subsequently, they add an additional condition, according to which 'An AI gained scientific understanding if it can transfer its understanding to a human expert'[25]. They then combine these two conditions into a test, which they describe as follows [25]:

> A human (the student) interacts with a teacher, either a human or an artificial scientist. The teacher's goal is to explain a scientific theory and its qualitative, characteristic consequences to the student. Another human (the referee) tests both the student and the teacher independently. If the referee cannot distinguish between the qualities of their non-trivial explanations in various contexts, we argue that the teacher has scientific understanding.

While promising, this approach may have a few issues. First, it equates teaching abilities with understanding, which is problematic if the student is simply a bad learner (despite the teacher's understanding). Second, the referee determines understanding by comparing the qualities of explanations. If both teacher and student lack understanding (whether because they simply lack explanations or because their explanations are incorrect), their explanations may be indistinguishable (by being equally wrong). According to the test, we should conclude the teacher has understanding. Third, the test is difficult to implement in practice due to vague parameters, such as the referee's inability to distinguish between non-trivial explanations in different contexts. The quality of explanations depends on explanatory aims and the variety of contexts is unclear, leading to different results depending on the chosen referee.

Despite its limitations, we think Krenn et al.'s test is valuable, and we propose a reformulation of it using our framework. Instead of relying on a referee, we propose to measure the student's score before and after interacting with a teacher to demonstrate an increase in scientific understanding (by the student). Additionally, we can test the teacher's understanding separately to distinguish between the teacher's own understanding and their ability to transfer that understanding to the student. We then suggest the reformulated test for scientific understanding transfer:

> The student takes an initial test, interacts with the teacher, and then takes a second test. While the second test should cover the same material or aspects, it should contain different questions to ensure the validity of the test. The extent to which the student's score increases on the second test is an indication of the teacher's ability to effectively teach phenomenon P to the student.

We view this reformulation as an improvement because it enables measuring the increase of scientific understanding in agents. This becomes particularly important when AI has developed new knowledge that needs to be conveyed to humans who lack that understanding. The reformulated test can be helpful in important cases where AI has developed new knowledge that needs to be conveyed to humans.

## Applications, limitations, and new scientific understanding

The usefulness of AI models, such as large language models (LLMs), in scientific contexts like hypothesis generation and information retrieval relies on aligning their scientific understanding with humans. The proposed framework for assessing and directing scientific understanding in AI agents has the potential to enhance the usefulness of AI models in scientific contexts. For instance, it can compare the performance of different AI agents in answering questions, as well as highlighting their strengths and limitations. Additionally, it can also aid in educational programs. By helping select relevant AI tools, this framework could be a valuable resource for students, serving as a pedagogical aid. Some AI models are already capable of performing above the level of a college student who has completed one semester of physics[60], highlighting the potential of these models as a valuable resource for students and researchers alike in the near future.

However, there are still open questions and challenges that need to be addressed. Establishing a threshold to determine sufficient understanding for an agent can prove to be complex, particularly when there may not be a consensus on the appropriate criteria. Similarly, in some testing modalities, experts might not always be available to check what the correct answer is, and for some questions we simply do not yet know the answer. However, this could open the possibility for new avenues of research, as asking these questions to QAMs might in some cases provide interesting answers that can trigger new avenues for research and stimulate hypothesis generation[30]. In such a case, if an agent is capable of answering questions for which humans do not have an answer yet, it may possess new scientific understanding.

To evaluate this *new scientific understanding*, a community can define w-questions whose answers are not yet known, similar to conjectures in mathematics that can be expected to be solved soon, where AI can tentatively provide unverified answers[61][62]. These lists of w-questions can measure progress in gaining new scientific knowledge and test forms of new scientific understanding. It is important to determine whether current LLMs/QAMs have new scientific understanding by asking questions where we might not know the answer. One can then employ our framework to act as a retroactive new scientific understanding test once these discoveries are confirmed or denied. This approach can encourage discovery and stimulate further research.

Finally, choosing the right prompt[63] (e.g., what context needs to be provided for each question) and evaluating vague outputs can be challenging. The Reinforcement Learning[64] used to fine tune certain models (optimizing for user experience) may steer in the direction of vague or incorrect answers which are detrimental to scientific research[51]. Addressing this issue may require AI models specifically designed for scientific aims, which could involve adjusting the training set[65] or prioritizing scientific understanding during training or fine tuning. Creating AI models tailored to scientific understanding has the potential to transform how we approach scientific exploration.

## Conclusion

This Roadmap presents a framework and testing methodology to assess agents' scientific understanding, a critical component in today's research landscape. It provides discussions and guidelines for communities to create their own scientific understanding tests, stressing their importance. The potential impact of this framework is far-reaching, as it can enhance the usefulness of AI, assess possible new scientific understanding encoded in machines, and aid in educational programs.

Future research directions include refining the methodology for creating tests and the concrete elaboration of tests which will form part of the benchmark for scientific understanding. Ultimately, scientific understanding tests are necessary to analyze, control, and harness the potential of AI in the context of scientific research.

[1] Clark, A., & Chalmers, D. The extended mind. *Analysis*, *58*(1), 7-19 (1998).
[2] Kuorikoski, J., & Ylikoski, P. External representations and scientific understanding. *Synthese*, *192*, 3817-3837(2015).
[3] Thiyagalingam, J., Shankar, M., Fox, G., & Hey, T. Scientific machine learning benchmarks. *Nature Reviews Physics*, *4*(6), 413-420 (2022).
[4] Li, Y., & Zhan, J. SAIBench: Benchmarking AI for science. *BenchCouncil Transactions on Benchmarks, Standards and Evaluations*, *2*(2), 100063 (2022).
[5] Levesque, H. J., Davis, E., & Morgenstern, L. The Winograd schema challenge. *KR*, *2012*, 13th (2012).
[6] Wang, A., Singh, A., Michael, J., Hill, F., Levy, O., & Bowman, S. R. GLUE: A multi-task benchmark and analysis platform for natural language understanding. *arXiv preprint arXiv:1804.07461*(2018).
[7] Srivastava, A., Rastogi, A., Rao, A., Shoeb, A. A. M., Abid, A., Fisch, A., et al. Beyond the imitation game: Quantifying and extrapolating the capabilities of language models. *arXiv preprint arXiv:2206.04615* (2022).
[8] Vaswani, A., Shazeer, N., Parmar, N., Uszkoreit, J., Jones, L., Gomez, A. N., et al. Attention is all you need. *Advances in neural information processing systems*, *30* (2017).
[9] Brown, T., Mann, B., Ryder, N., Subbiah, M., Kaplan, J. D., et al. Language models are few-shot learners. *Advances in neural information processing systems*, *33*, 1877-1901 (2020).
[10] Allam, A. M. N., & Haggag, M. H. The question answering systems: A survey. *International Journal of Research and Reviews in Information Sciences (IJRRIS)*, *2*(3) (2012).

[11] De Regt, H. W., Leonelli, S., & Eigner, K. (Eds.). *Scientific understanding: Philosophical perspectives*. University of Pittsburgh Press (2009).

[12] De Regt, H. W. *Understanding scientific understanding*. Oxford University Press (2017).

[13] Dellsén, F. Beyond explanation: Understanding as dependency modelling. *The British Journal for the Philosophy of Science* (2020).

[14] Wilkenfeld, D. A. Understanding as representation manipulability. *Synthese*, *190*, 997-1016 (2013).

[15] Searle, J. R. Minds, brains, and programs. *Behavioral and Brain Sciences*, *3*(3), 417-424 (1980).

[16] Johnson-Laird, P. N. Mental models and human reasoning. *Proceedings of the National Academy of Sciences*, *107*(43), 18243-18250(2010).

[17] Nersessian, N. J. How do scientists think? Capturing the dynamics of conceptual change in science. *Cognitive Models of Science*, *15*, 3-44(1992).

[18] Marcus, G. Deep learning: A critical appraisal. *arXiv preprint arXiv:1801.00631* (2018).

[19] Chollet, F. The limitations of deep learning. *Deep learning with Python*. (2017).

[20] Tamir, M., & Shech, E. Machine understanding and deep learning representation. *Synthese*, *201*(2), 51(2023).

[21] Rozenblit, L., & Keil, F. The misunderstood limits of folk science: An illusion of explanatory depth. *Cognitive science*, *26*(5), 521-562 (2002).

[22] De Regt, H. W., & Dieks, D. A contextual approach to scientific understanding. *Synthese*, *144*, 137-170 (2005).

[23] Turing, A. M. Computing machinery and intelligence. *Mind*, 49(236), 433-460 (1950).

[24] Oppy, G. & Dowe, D. The Turing Test. *The Stanford Encyclopedia of Philosophy*, Edward N. Zalta (ed.) (Winter 2021 Edition)

[25] Krenn, M., Pollice, R., Guo, S. Y., Aldeghi, M., Cervera-Lierta, A., Friederich, P., dos Passos Gomes, G., Häse, F., Jinich, A., Nigam, A., Yao, Z., & Aspuru-Guzik, A. On scientific understanding with artificial intelligence. *Nature Reviews Physics*, *4*(12), 761-769 (2022).

[26] Baumberger, C., Beisbart, C., and Brun, G. What is understanding? An overview of recent debates in epistemology and philosophy of science. In *Explaining understanding: new perspectives from epistemology and philosophy of science*. Eds. Grimm, S. R., Baumberger, C., and Ammon S. Routledge (2017) (pp.1-34).

[27] Grimm, S. R. Understanding. *The Stanford Encyclopedia of Philosophy*. Edward N. Zalta (ed.) (Summer 2021 Edition).

[28] Floridi, L. AI as Agency without Intelligence: On ChatGPT, large language models, and other generative models. *Philosophy & Technology*, *36*(1), 15 (2023).

[29] Bender, E. M., Gebru, T., McMillan-Major, A., & Shmitchell, S. On the Dangers of Stochastic Parrots: Can Language Models Be Too Big? 🦜. In *Proceedings of the 2021 ACM conference on fairness, accountability, and transparency* (pp. 610-623) (2021, March).

[30] Bubeck, S., Chandrasekaran, V., Eldan, R., Gehrke, J., Horvitz, E., Kamar, et al. Sparks of Artificial General Intelligence: Early experiments with GPT-4. *arXiv preprint arXiv:2303.12712*(2023).

[31] De Regt, H. W. *Understanding scientific understanding*. Oxford University Press (2017, p.92).

[32] De Regt, H. W. *Understanding scientific understanding*. Oxford University Press (2017, p.102).

[33] Woodward, J. and Ross, L. Scientific Explanation. *The Stanford Encyclopedia of Philosophy*. Edward N. Zalta (ed.) (Summer 2021 Edition)

[34] Woodward, J. *Making Things Happen: A theory of causal explanation*. Oxford University Press (2003).

[35] Hitchcock, C., & Woodward, J. Explanatory generalizations, part II: Plumbing explanatory depth. *Noûs*, *37*(2), 181-199 (2003).

[36] Weslake, B. Explanatory depth. *Philosophy of Science*, *77*(2), 273-294 (2010).

[37] Kuorikoski, J., & Ylikoski, P. External representations and scientific understanding. *Synthese*, *192*, 3817-3837 (2015).

[38] Belnap, N. D., & Steel, T. B. The logic of questions and answers (1976).

[39] Cross, C., and Roelofsen, F., "Questions", *The Stanford Encyclopedia of Philosophy*, Edward N. Zalta (ed.) (Summer 2022 Edition)

[40] Cross, C. B. Explanation and the Theory of Questions. *Erkenntnis*, *34*(2), 237-260 (1991).

[41] Bromberger, S. *Why-questions.* In *Mind and Cosmos: Essays in Contemporary Science and Philosophy,* ed. Colodny, R.G. University of Pittsburgh Press (pp. 86-111) (1966).

[42] Hempel, C. G., & Oppenheim, P. Studies in the Logic of Explanation. *Philosophy of science*, *15*(2), 135-175(1948).

[43] Van Fraassen, B. C. *The Scientific Image*. Oxford University Press (1980).

[44] Weber, E., & Lefevere, M. Unification, the answer to resemblance questions. *Synthese*, *194*, 3501-3521(2017).

[45] Weber, E., van Eck, D., & Mennes, J. On the structure and epistemic value of function ascriptions in biology and engineering sciences. *Foundations of Science*, *24*, 559-581(2019).

[46] Ylikoski, P., & Kuorikoski, J. Dissecting explanatory power. *Philosophical studies*, *148*, 201-219(2010).

[47] Barman, K. G. Procedure for assessing the quality of explanations in failure analysis. *AI EDAM*, *36* (2022).

[48] Kuorikoski, J., & Ylikoski, P. External representations and scientific understanding. *Synthese*, *192*, 3817-3837 (2015).

[49] Pearl, J. Causal inference in statistics: An overview. *Statistics Surveys Vol. 3,* 96–146 (2009).

[50] Halpern, J. Y. *Actual Causality*. Cambridge, MA: MIT Press (2016).

[51] Perez, E., Ringer, S., Lukošiūtė, K., Nguyen, K., Chen, E., Heiner, S., ... & Kaplan, J. Discovering Language Model Behaviors with Model-Written Evaluations. *arXiv preprint arXiv:2212.09251* (2022).

[52] Du, X., Shao, J., & Cardie, C. Learning to ask: Neural question generation for reading comprehension. *arXiv preprint arXiv:1705.00106* (2017).

[53] Rao, S., & Daumé III, H. Learning to ask good questions: Ranking clarification questions using neural expected value of perfect information. *arXiv preprint arXiv:1805.04655* (2018).

[54] Mintzes, J. J., Wandersee, J. H., & Novak, J. D. (Eds.). *Assessing Science Understanding: A human constructivist view*. Academic Press (2005).

[55] Schleicher, A. *Measuring Student Knowledge and Skills: A New Framework for Assessment*. Organisation for Economic Co-Operation and Development, Paris, France (1999).

[56] Franzen, M. Assessing Student Understanding in Science. *Science and Children*, *47*(9), 79 (2010).

[57] Brookhart, S. M. How to Create and Use Rubrics for Formative Assessment and Grading. ASCD (2013).

[58] https://openai.com/blog/chatgpt

[59] https://futureoflife.org/open-letter/pause-giant-ai-experiments/?ftag=YHF4eb9d17

[60] West, C. G. AI and the FCI: Can ChatGPT Project an Understanding of Introductory Physics? *arXiv preprint arXiv:2303.01067* (2023).

[61] Ganesalingam, M., & Gowers, W. T. A fully automatic theorem prover with human-style output. *Journal of Automated Reasoning*, *58*, 253-291(2017).

[62] https://ai.facebook.com/blog/ai-math-theorem-proving/

[63] Reynolds, L., & McDonell, K. Prompt programming for large language models: Beyond the few-shot paradigm. In *Extended Abstracts of the 2021 CHI Conference on Human Factors in Computing Systems* (pp. 1-7) (2021, May).

[64] Sutton, R. S., & Barto, A. G. *Reinforcement learning: An introduction*. MIT press (2018).

[65] Taylor, R., Kardas, M., Cucurull, G., Scialom, T., Hartshorn, A., Saravia, E., ... & Stojnic, R. Galactica: A large language model for science. *arXiv preprint arXiv:2211.09085* (2022).